\pdfoutput=1
\documentclass[11pt]{article}

\usepackage{EMNLP2023}

\usepackage{times}
\usepackage{latexsym}

\usepackage[T1]{fontenc}
\usepackage[utf8]{inputenc}

\usepackage{microtype}

\usepackage{inconsolata}

\usepackage{graphicx}
\usepackage{algorithm}
\usepackage{algorithmic}
\usepackage{multirow}
\usepackage{booktabs}
\usepackage{amsmath}
\usepackage{amssymb}
\usepackage{amsfonts}
\usepackage{cleveref}
\usepackage{mathrsfs}
\usepackage{comment}
\DeclareMathOperator*{\argmax}{\text{argmax}} % Define the argmax operator

\title{Multilingual Knowledge Graph Completion via Efficient Multilingual Knowledge Sharing}

\author{
Cunli Mao$^{1,2}$, 
Xiaofei Gao$^{1,2}$, 
Ran Song$^{1,2}$\thanks{Corresponding author}, 
Shizhu He $^{3,4}$ \\ 
{\bf Shengxiang Gao$^{1,2}$,
Kang Liu $^{3,4}$, 
Zhengtao Yu$^{1,2}$ } \\
$^{1}$Faculty of Information Engineering and Automation, \\
Kunming University of Science and Technology, Kunming, China \\
$^{2}$Yunnan Key Laboratory of Artificial Intelligence, Kunming, China \\
$^{3}$The Key Laboratory of Cognition and Decision Intelligence for Complex Systems, \\
Institute of Automation, Chinese Academy of Sciences, Beijing, China \\
$^{4}$School of Artificial Intelligence, University of Chinese Academy of Science, Beijing, China \\
\{maocunli,xiaofeigao\_g,song\_ransr\}@163.com, \{shizhu.he,kliu\}@nlpr.ia.ac.cn, \\
\{gaoshengxiang.yn,ztyu\}@hotmail.com
}

\begin{document}
\maketitle
\begin{abstract}
Large language models (LLMs) based Multilingual Knowledge Graph Completion (MKGC) aim to predict missing facts by leveraging LLMs' multilingual understanding capabilities, improving the completeness of multilingual knowledge graphs (KGs).
However, existing MKGC research underutilizes the multilingual capabilities of LLMs and ignores the shareability of cross-lingual knowledge.
In this paper, we propose a novel MKGC framework that leverages multilingual shared knowledge to significantly enhance performance through two components: Knowledge-level Grouped Mixture of Experts (KL-GMoE) and Iterative Entity Reranking (IER).
KL-GMoE efficiently models shared knowledge, while IER significantly enhances its utilization.
To evaluate our framework, we constructed a mKG dataset containing 5 languages and conducted comprehensive comparative experiments with existing state-of-the-art (SOTA) MKGC method.
The experimental results demonstrate that our framework achieves improvements of 5.47\%, 3.27\%, and 1.01\% in the Hits@1, Hits@3, and Hits@10 metrics, respectively, compared with SOTA MKGC method.
Further experimental analysis revealed the properties of knowledge sharing in settings of unseen and unbalanced languages.
We have released the dataset and code for our work on \url{https://github.com/gaoxiaofei07/KL-GMoE}.
\end{abstract}

\begin{figure}[t]
  \centering
  \includegraphics[width=\columnwidth]{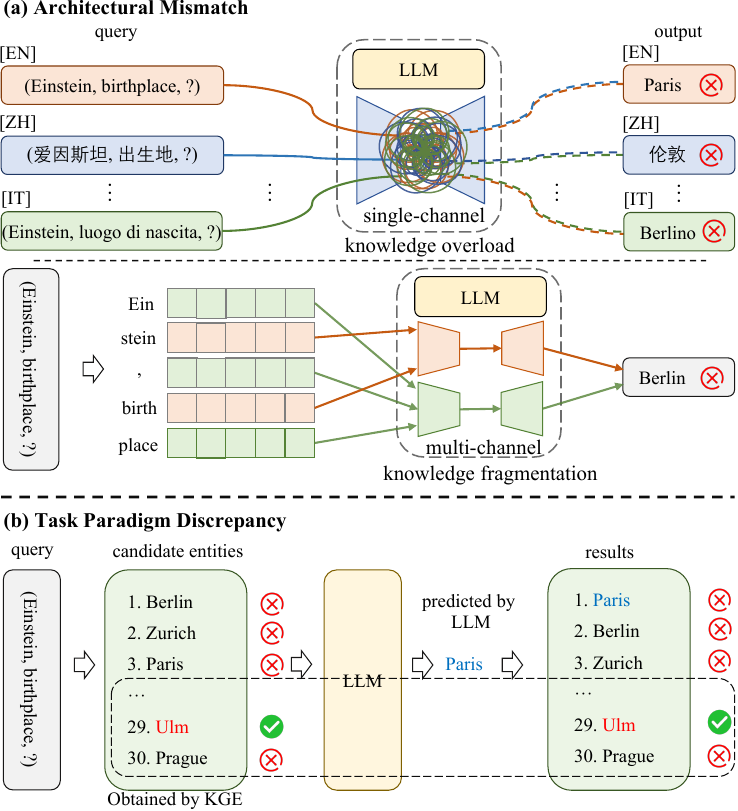}
  \caption{
      This figure depicts the problems encountered when applying LLMs directly to the MKGC task.
      (a) illustrates that existing PEFT is not suitable for the MKGC task.
      (b) Indicates a discrepancy between the task paradigms LLMs excel at and MKGC tasks.
  }
  \label{fig:introduction_pdf}
\end{figure}

\section{Introduction}
Knowledge Graphs (KGs) \citep{weikum2021knowledge} are structured semantic knowledge bases designed to represent and organize knowledge about the real world. 
Most KGs possess multilingual characteristics, including \textit{Wikidata} \citep{vrandevcic2014wikidata} and \textit{DBpedia} \citep{lehmann2015dbpedia}.
Existing multilingual KGs are often incomplete, which limits their effectiveness in practical applications \citep{9416312}.
Multilingual knowledge graph completion (MKGC) aims to leverage known multilingual knowledge to complete missing triples and improve the completeness of the KGs.

Studies have focused on embedding-based methods \citep{ge2024knowledge} for MKGC, mapping entities and relations into a low-dimensional vector space to achieve completion. 
Recent advances in language models have shifted MKGC research toward generation-based approaches \citep{chen-etal-2022-knowledge, saxena-etal-2022-sequence} that reformulate KG completion as text generation. 
Methods \citep{song-etal-2023-multilingual, zhou-etal-2022-prix} employ a single pretrained language model (PLM) to consolidate multilingual knowledge within a unified semantic space, achieving superior performance in MKGC. 
Furthermore, recent work like DIFT \citep{liu2024finetuning} explores task adaptation through LLMs fine-tuning, achieving strong performance on monolingual KGC tasks. 
Modern LLMs, pretrained on diverse corpora, inherently possess multilingual capabilities \citep{huang2024survey}, enabling the representation and knowledge sharing across languages \citep{hu-etal-2025-large-language} within a unified model.
Crucially, this capacity for internal multilingual knowledge sharing is vital for MKGC, offering substantial potential to enhance completion performance.
Motivated by this recognized potential, our research investigates effective methods for harnessing LLMs' inherent capabilities to improve MKGC.

However, directly applying LLMs to MKGC presents several challenges, primarily stemming from two key aspects: model architecture and task paradigm.
1) \textbf{Architectural Mismatch}:
Existing Parameter-Efficient Fine-Tuning (PEFT) methods \citep{han2024parameter} for LLMs are mainly designed for text-centric tasks and exhibit significant gaps when applied to knowledge-level tasks.
Specifically, single-channel methods struggle with the complex multilingual nature of KGs.
As shown in Figure~\ref{fig:introduction_pdf}(a) top, processing numerous multilingual queries through a single channel often results in knowledge overload.
This overload impacts the model's ability to understand similar knowledge across languages, leading to incorrect predictions. 
 For example, queries in English, Chinese, and Italian concerning \textit{Einstein's birthplace} all yield incorrect results.
Conversely, multi-channel methods tend to disrupt the atomicity of knowledge, thereby causing knowledge fragmentation. 
As shown in Figure~\ref{fig:introduction_pdf}(a) bottom, query tokens are processed by disparate channels.
Such fragmented processing consequently leads to incorrect entity predictions.
2) \textbf{Task Paradigm Discrepancy}:
The MKGC task involves entity ranking, which presents a discrepancy with the text generation paradigm.
As shown in Figure~\ref{fig:introduction_pdf}(b), for the query \textit{(Einstein, birthplace, ?)}, the LLM erroneously predicted \textit{Paris} as the answer.
This selection failed to improve the ranking of the correct entity \textit{Ulm}.

To address the 1) \textbf{Architectural Mismatch}, specifically knowledge overload, we propose increasing the number of dedicated knowledge channels.
This allows each channel to focus on processing semantically similar information, thereby enhancing the LLM's capacity to understand and leverage cross-lingual shared knowledge. 
Concurrently, by enabling each channel to independently process complete knowledge, we can effectively mitigate knowledge fragmentation and facilitate the model's comprehensive understanding of multilingual information.
To resolve the 2) \textbf{Task Paradigm Discrepancy}, we propose adjusting the LLM's training objective to enable it to iteratively refine the ranking of multiple entities. 
This approach aims to enhance the ranking of correct entities by increasing the frequency with which the LLM utilizes cross-lingual shared knowledge.

In this paper, we propose a novel framework for effectively leveraging multilingual shared knowledge to enhance the performance of MKGC.
This proposed framework comprises two synergistic components: Knowledge-level Grouped Mixture of Experts (KL-GMoE) and Iterative Entity Reranking (IER).
KL-GMoE introduces a knowledge-level expert routing mechanism and a group-based Mixture-of-Experts (MoE) architecture. 
This design aims to mitigate knowledge fragmentation while substantially enhancing LLMs' capacity to capture cross-lingual shared knowledge.
IER modifies both the training objective and the decoding strategy of LLMs. 
This enables the models to significantly improve their leveraging of multilingual shared knowledge through multiple iterative refinements.
The experimental results demonstrate that our framework achieves improvements of 5.47\%, 3.27\%, and 1.01\% in the Hits@1, Hits@3, and Hits@10 metrics, respectively, compared with SOTA MKGC method.
Further experimental analysis revealed the properties of knowledge sharing in settings of unseen and unbalanced languages.

In summary, our contributions are as follows: 
\begin{itemize}
    \itemsep0em % Adjust this value as needed (e.g., 0.5em, -0.2em)
    \item We propose KL-GMoE to address the model architecture mismatch, efficiently modeling shared knowledge.
    \item We propose IER to address the discrepancy in the task paradigm, enhancing the utilization of shared knowledge.
    \item Experiments show that our framework significantly outperforms the SOTA MKGC method, with average improvements of 5.47\%, 3.27\%, and 1.01\% in Hits@1, Hits@3, and Hits@10.
\end{itemize}

\section{Datasets}

\subsection{Dataset Construction}
We utilize \textit{Wikidata5M} \citep{wang-etal-2021-kepler} as the foundational seed library, which is a million-scale English KG dataset integrating \textit{Wikidata} and \textit{Wikipedia}.
Based on this, we further expanded the dataset to include French, Italian, Chinese, and Japanese, by collecting data from \textit{Wikidata}.
As shown in Table~\ref{mkgc-dataset}, we present statistics on the number of entities, relations, training, validation and testing triples.
The KG contains 351,299 entities and 2,264 relations, with the total number of triples exceeding 3 million.

Based on the characteristics of multilingual knowledge distribution, the knowledge across different languages is not entirely aligned but exhibits certain linguistic specificity \citep{SONG2025130979}.
This asymmetry of knowledge across languages indicates that some knowledge is confined to specific languages.
Therefore, the dataset we constructed follows the natural distribution patterns of knowledge.
Some knowledge is shared across multiple languages, reflecting the similarities between languages.
Other knowledge is unique to each language, reflecting the distinctive characteristics of each language.

\begin{table}[t]
    \centering
    \small 
    \setlength{\tabcolsep}{1.1pt} 
        \begin{tabular}{cccccc}
        \hline
            \small \textbf{Language} & \small \textbf{Entity} & \small \textbf{Relation} & \small \textbf{Training} & \small \textbf{Validation} & \small \textbf{Testing}\\
            \hline
            EN & 86,539 & 512 & 708,267 & 49,782 & 49,777 \\
            FR & 89,754 & 478 & 839,623 & 49,908 & 30,000 \\
            IT & 65,434 & 445 & 613,014 & 49,883 & 20,000 \\
            JA & 46,294 & 432 & 321,237 & 49,939 & 10,000 \\
            ZH & 63,278 & 397 & 546,626 & 49,969 & 10,000 \\
            \midrule
            SUM & 351,299 & 2,264 & 3,028,767 & 249,481 & 119,777 \\
        \hline
        \end{tabular}
    \caption{\label{mkgc-dataset} Statistics of the multilingual knowledge graph completion dataset.}
\end{table}

\subsection{Prompt Construction}
We adopt the prompt construction method proposed by DIFT \citep{liu2024finetuning}.
Since the embedding-based model has learned the training data, it tends to rank the correct entity at the first in the candidate entities for most training facts.
Constructing the prompt using these ranked candidates may cause LLMs to develop a bias toward selecting the first entity as the answer.
Therefore, we partition a subset from the validation set to construct prompts, which are utilized as training data during the fine-tuning phase of the LLM.

For the query \( q = (h, r, ?) \), the constructed Prompt \( \text{P} \) consists of four parts: Query \( Q \), Description \( D \), Neighbor facts \( N \), and Candidate entities \( M_c \).
This can be represented as:
\begin{align}
    P(q) &= [Q; D; N; M_c].
\end{align}

Description provides specific descriptive information about entity \( h \), enabling the model to comprehend the entity's meaning more accurately.
Neighbor facts are triples that include the entity \( h \), and these triples are randomly sampled from the Knowledge Graph Embedding (KGE) model's training data.
These neighboring facts are intended to enhance the LLM's comprehension of the entity \( h \). 
Candidate entities \( M_c  = [e_1, e_2, \dots, e_m] \) are composed of the top-\( m \) entities selected from the ranking results generated by the KGE model \citep{bordes2013translating}.
To enable LLMs to adapt to the task paradigm of MKGC, we processed the number of entities \( m \) in the \( M_c \) during training.
The specific processing method is detailed in Section~\ref{sec:ier-method}.
We provide specific prompt examples in Appendix~\ref{Appendix:prompt-example}.

\definecolor{myred}{RGB}{226, 80, 90}

\begin{figure*}[t]
  \centering
  \includegraphics[width=\textwidth]{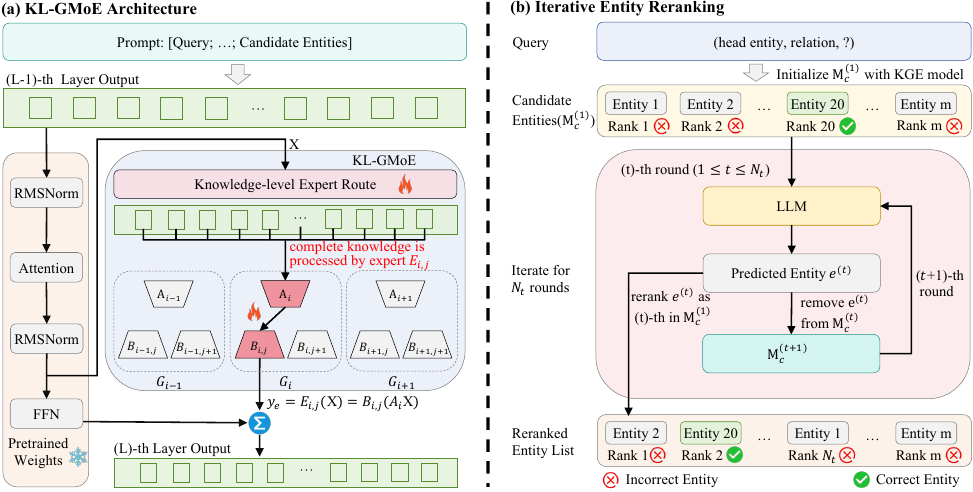}
  \caption{
    The figure illustrates our proposed framework.
    Figure (a) depicts the architecture and workflow of the KL-GMoE, where the matrices \( A_{i} \) and \( B_{i,j} \) highlighted in \textcolor{myred}{red} represent the currently activated expert.
    Figure (b) illustrates the workflow of the IER method.
    After $N_t$ iterations, we can obtain a reranked list of entities.
  }
  \label{fig:method_pdf}
\end{figure*}

\section{Methodology}

\subsection{Task Definition}
In this paper, we integrate KGE model with LLM to perform the MKGC task.
First, for a query \( q = (h, r, ? ) \), we use the KGE model to obtain the top-\(m\) ranked entities, which form the candidate entities \( M_c \).
Next, we leverage LLMs to select the optimal entity from the candidate entities \( M_c \) to complete the query \( q \).
The completion process can be formulated as follows:
\begin{align}
    \hat{e} = \underset{e_i \in M_c}{\text{argmax}} \, P(e_i \mid h, r, M_c),
\end{align}
where \( \hat{e} \) denotes the optimal entity for completing the query \( q \), and \( P(e_i \mid h, r, M_c) \) represents the probability of selecting entity \( e_i \) given the head entity \( h \), relation \( r \) and candidate entities \( M_c \).

\subsection{KL-GMoE Architecture}
KL-GMoE is specifically tailored for MKGC task.
This architecture is designed with multiple expert groups to alleviate the knowledge overload caused by single-channel and enhance the ability of LLMs to capture shared knowledge.
Furthermore, KL-GMoE employs a knowledge-level expert routing mechanism to ensure that each sample is processed by a specific expert, rather than involving all experts collectively.
As shown in Figure~\ref{fig:method_pdf}(a), for each sample processed by KL-GMoE, only the matrix \( A_i \) and one matrix \( B_{i,j}\) from the expert group \( G_i \) are activated.
The KL-GMoE is applied exclusively to the Feed-Forward Network (FFN) layer of the LLM.
Specifically, the matrix operations in the FFN layer during forward propagation can be represented as follows:
\begin{equation} \label{eq:ffn}
    \mathbf{y} = \mathbf{W}_0 \mathbf{X} + \mathbf{y}_{e},
\end{equation}
where \( \mathbf{W}_0 \in \mathbb{R}^{\text{dout} \times \text{din}} \) represents the original FFN layer parameter matrix, which is frozen during training.
\( \mathbf{X} = \mathbf{[x_h:x_r:x_t]} \) represents the input of the FFN layer.
\( \mathbf{y}_{e} \) represents the output calculated by KL-GMoE based on the input \( \mathbf{X} \) .

The design of KL-GMoE is inspired by the asymmetric fine-tuning architecture proposed in HydraLoRA \citep{tian2024hydralora}. 
We adopted a grouped MoE design architecture, where each group can be represented as follows:
\begin{equation} \label{eq:ffn_expert}
    G_{i} = (A_{i}, \{B_{{i},{j}} \mid {j} \in \{1, 2, \dots, N_b\}\}),
\end{equation}
where \( i \in \{1, 2, \dots, N_g\} \), \(N_g\) denotes the total number of expert groups.
\(N_b\) is the number of B matrices in group \( G_{i} \) .
Within each group \( G_{i} \), the pairing of the \( A_{i} \) matrix with a \( B_{{i},{j}} \) matrix is considered an expert \( E_{{i},{j}} = (A_{i}, B_{{i},{j}}) \).
The \( A_{i} \) matrix is designed to capture a category of similar knowledge.
The different \( B_{{i},{j}} \) matrices within the group \( G_{i} \) are regarded as modules that capture subtle differences in this category of knowledge.
This design aims to enhance LLMs' ability to capture shared knowledge across multiple languages.

Simultaneously, our proposed knowledge-level expert routing mechanism includes three different routes: \( \mathbf{R}_g \), \( \mathbf{R}_k \) and \( \mathbf{R}_l \). 
First, an expert group is selected based on \( \mathbf{R}_g \).
Within this group, a specific expert is then determined by combining \( \mathbf{R}_k \) and \( \mathbf{R}_l \).
The following describes the process of selecting a specific expert based on these three routes.

\(\mathbf{R}_g\) is the group routing selection module that determines which expert group processes the \( \mathbf{X} \).
The group selection is formulated as follows:
\begin{equation}
    \label{eq:group_routing}
    \begin{aligned}
        G_i &= \argmax \limits_{i \in \{1, 2, \dots, N_g\}} \left( \mathbf{R}_g(\mathbf{X}) \right) \\
        &= \argmax \limits_{i \in \{1, 2, \dots, N_g\}} \left( \sum_{m \in \{h, r, t\}} \text{Softmax}(\mathbf{W}_g \mathbf{x}_m) \right),
    \end{aligned}
\end{equation}
where \( \mathbf{W}_g \in \mathbb{R}^{N_g \times \text{din}} \) is the routing matrix for group selection. 
\( {G_i} \) represents the expert group selected to process \( \mathbf{X} \).

\(\mathbf{R}_{k}\) and \(\mathbf{R}_{l}\) represent expert routing selection modules that operate within the group \( G_{i} \).
These modules comprehensively considers the input \( \mathbf{X} \) and the output from the \( A_{i} \) matrix to perform expert selection.
Specifically, \(\mathbf{R}_{k}\) generates expert selection scores \( \mathbf{S}_k \) based on \( \mathbf{X} \).
The formula for calculating \( \mathbf{S}_k \) is as follows:
\begin{equation} \label{eq:expert_routing_k}
    \begin{aligned}
        \mathbf{S}_k = \mathbf{R}_{k}(\mathbf{X}) =  \sum_{m \in \{h, r, t\}} \text{Softmax}(\mathbf{W}_k \mathbf{x}_m),
    \end{aligned}
\end{equation}
where \( \mathbf{S}_k \in \mathbb{R}^{N_b} \), and \( \mathbf{W}_k \in \mathbb{R}^{N_b \times \text{din}} \) is the routing matrix that receives \( \mathbf{X} \) as input.
\( \mathbf{R}_{l} \) generates expert selection scores \( \mathbf{S}_l \) based on the output \( A_{i}\mathbf{X} \) of matrix \( A_{i} \).
The calculation of \( \mathbf{S}_l \) can be expressed as follows:
\begin{equation} \label{eq:expert_routing_l}
    \begin{aligned}
        \mathbf{S}_l = \mathbf{R}_{l}(\mathbf{X}) = \sum_{m \in \{h, r, t\}} \text{Softmax}(\mathbf{W}_l (A_{i}\mathbf{x}_m)),
    \end{aligned}
\end{equation}
where \( \mathbf{S}_l \in \mathbb{R}^{N_b} \), and \( \mathbf{W}_l \in \mathbb{R}^{N_b \times \text{r}} \) is the routing matrix that receives \( A_{i}\mathbf{X} \) as input.
\( r \) represents the size of the rank in LoRA \citep{hu2022lora}.
Then, select the matrix \( B_{i, j} \) from group \( G_i \) based on the scores of \( \mathbf{S}_k \) and \( \mathbf{S}_l \) to process \( \mathbf{X} \):
\begin{equation} \label{eq:expert_selection}
    B_{i, j} = \argmax \limits_{j \in \{1, 2, \dots, N_b\}} \left( \mathbf{S}_k + \mathbf{S}_l \right).
\end{equation}
Finally, we determine that the expert \( E_{i,j} = (A_{i}, B_{i, j}) \) processes  \( \mathbf{X} \) based on the knowledge-level expert routing mechanism.

After determining the expert \( E_{i,j} \), the output of KL-GMoE is expressed as follows:
\begin{equation} \label{eq:expert_contribution}
    \mathbf{y}_{e} = E_{i,j}(\mathbf{X}) = B_{i, j} ( A_{i} \mathbf{X} ).
\end{equation}
Then, the expert output \( \mathbf{y}_{e} \) is added to the original FFN output, as shown in Equation~\ref{eq:ffn}.

\subsection{Iterative Entity Reranking}\label{sec:ier-method}
We propose a method called Iterative Entity Reranking (IER), aimed at enhancing LLMs' utilization of cross-lingual shared knowledge.
As shown in Figure \ref{fig:method_pdf}(b), the IER method fully leverages shared knowledge through multiple iterations, significantly improving the accuracy of correct entity ranking.
IER adjusts the training task and decoding strategy of LLMs.
In the training phase, we randomly set the number of candidate entities \( m \) to a variable value, to train the LLM to be capable of iteratively adjusting the ranking of multiple entities.
In the decoding stage, IER allows the LLMs to perform multiple rounds of entity prediction to adjust the ranking of multiple entities.

For the query \( q = (h, r, ?) \), the initial set of candidate entities is generated by the KGE model and denoted as \( M_c^{(1)} = [e_1, e_2, \dots, e_m] \). 
The list of entities to be sorted is initialized as \( L^{(1)} = M_c^{(1)} \).
The LLM performs \( N_t \) rounds of entity prediction. 
In the \( t \)-th round, where \(  t \in \{1, 2, \dots, N_t \}\), the entity prediction operation can be expressed as follows:
\begin{equation}
    e^{(t)} = \underset{e_i \in M_c^{(t)}}{\text{argmax}} \, P(e_i \mid h, r, M_c^{(t)}),
\end{equation}
where \(  M_c^{(t)} \) represents the candidate entity set in round \( t \). \( e^{(t)} \) is the entity predicted by the LLM from \( M_c^{(t)} \).
Then,  we update \( M^{(t)} \) to obtain \(M^{(t+1)}\) for the next iteration:
\begin{equation}
    M_c^{(t+1)} = M_c^{(t)} \setminus \{e^{(t)}\},
\end{equation}
where \( M_c^{(t)} \setminus \{e^{(t)}\} \) denotes removing the entity \( e^{(t)} \) from \( M_c^{(t)} \).
Finally,we update the ranking of entity \( e^{(t)} \) in \( L^{(t)} \):
\begin{equation}
     L^{(t+1)} = \textbf{Insert}(L^{(t)} \setminus \{e^{(t)}\}, t, e^{(t)}),
\end{equation}
where \( \textbf{Insert}(L^{(t)} \setminus \{e^{(t)}\}, t, e^{(t)}) \) denotes first removing \( e^{(t)} \) from \( L^{(t)} \), and then inserting \( e^{(t)} \) into the \( t \)-th position of \( L^{(t)} \).
After iterating for \( N_t \) rounds, we obtain the final ranked list of entities \( L^{(N_t+1)} \).
The implementation of IER is detailed in Appendix~\ref{Appendix:ier-alg}.

\begin{table}[!ht]
    \centering
    \small
    \setlength{\tabcolsep}{2.5pt}
    \renewcommand{\arraystretch}{1.06}
    \begin{tabular}{p{0.2cm}ccccccc}
        \toprule
         & MODEL & EN & FR & IT & JA & ZH & AVG \\
        \midrule
        \multirow{10}{*}{\shortstack{H \\ @ \\ 1}}
            & TransE & 8.52 & 9.07 & 9.36 & 8.00 & 11.77 & 9.34 \\
            & Analogy & 13.40 & 15.81 & 14.58 & 15.11 & 6.43 & 13.07 \\
            & ComplEx & 10.92 & 11.75 & 11.49 & 13.95 & 17.79 & 13.18 \\
            & Distmult & 6.89 & 7.73 & 7.93 & 8.16 & 5.66 & 7.27  \\
            & RotatE & 24.08 & 24.61 & 25.57 & 29.49 & 31.36 & 27.02  \\
            & HAKE & 31.64 & 32.92 & 30.99 & 35.53 & 52.24 & 36.66  \\
            \cline{2-8}
            & ICL & 1.79 & 1.07 & 1.26 & 1.93 & 2.27 & 1.66 \\
            & GC-PLM & 33.37 & 32.51 & 30.38 & 36.65 & 49.13 &  36.41\\
            & DIFT (\textit{Single}) & 36.05 & 35.75 & 34.22 & 38.31 & 56.65 & 40.19 \\
            \cline{2-8}
            & Ours & \textbf{36.50} & \textbf{36.72} & \textbf{35.93} & \textbf{41.60} & \textbf{58.63} & \textbf{41.88} \\
        \midrule
        \multirow{10}{*}{\shortstack{H \\ @ \\ 3}} 
            & TransE & 37.02 & 39.17 & 37.57 & 44.59 & 60.51 & 43.78 \\
            & Analogy & 28.39 & 30.45 & 29.56 & 35.84 & 19.61 & 28.77  \\
            & ComplEx & 23.05 & 23.92 & 23.05 & 29.68 & 41.12 & 28.17 \\
            & Distmult & 14.59 & 14.82 & 15.79 & 19.16 & 17.94 & 16.46 \\
            & RotatE & 40.73 & 42.13 & 41.68 & 49.78 & 62.36 & 47.34  \\
            & HAKE & 43.30 & 43.27 & 41.88 & 47.52 & 63.22 & 47.84  \\
            \cline{2-8}
            & ICL & 34.99 & 37.09 & 35.33 & 42.84 & 59.15 & 41.88 \\
            & GC-PLM & 40.99 & 42.21 & 40.47 & 50.95 & 63.67 & 47.66 \\
            & DIFT (\textit{Single}) & 42.21 & 42.28 & 40.50 & 47.83 & 64.50 & 47.46 \\
            \cline{2-8}
            & Ours  & \textbf{46.25} & \textbf{45.30} & \textbf{44.22} & \textbf{51.97} & \textbf{66.93} & \textbf{50.93} \\
        \midrule
        \multirow{10}{*}{\shortstack{H \\ @ \\ 10}}
            & TransE  & 50.25 & 51.23 & 49.60 & 58.10 & 71.80 & 56.19 \\
            & Analogy & 39.17 & 41.73 & 40.30 & 48.78 & 64.45 & 46.89 \\
            & ComplEx & 34.84 & 37.51 & 35.70 & 45.02 & 59.97 & 42.61  \\
            & Distmult & 26.74 & 26.78 & 27.41 & 36.04 & 50.25 & 33.44  \\
            & RotatE & 52.66 & 53.17 & 51.50 & \textbf{61.68} & \textbf{74.58} & 58.72  \\
            & HAKE & 53.37 & 52.06 & 50.49 & 57.85 & 70.04 & 56.76  \\
            \cline{2-8}
            & ICL & 49.99 & 51.05 & 49.30 & 57.99 & 71.71 & 56.01 \\
            & GC-PLM & 52.76 & 52.81 & 51.76 & 59.53 & 71.98 & 57.77 \\
            & DIFT (\textit{Single}) & 52.48 & 52.35 & 50.30 & 58.74 & 72.08 & 57.19 \\
            \cline{2-8}
            & Ours  & \textbf{54.71} & \textbf{53.45} & \textbf{52.31} & 60.85 & 72.56 & \textbf{58.78} \\
        \midrule
        \multirow{10}{*}{\shortstack{M \\ R \\ R}}
            & TransE  & 24.85 & 25.97 & 25.46 & 28.38 & 37.51 & 28.43 \\
            & Analogy & 22.85 & 25.01 & 23.82 & 27.44 & 21.04 & 24.03 \\
            & ComplEx & 19.18 & 20.29 & 19.56 & 24.30 & 32.12 & 23.09  \\
            & Distmult & 13.33 & 13.80 & 14.28 & 16.86 & 16.69 & 14.99  \\
            & RotatE & 34.63 & 35.39 & 35.49 & 41.75 & 48.74 & 39.20 \\
            & HAKE & 39.41 & 39.80 & 38.10 & 43.36 & 59.04 & 43.94  \\
            \cline{2-8}
            & ICL & 20.11 & 20.70 & 19.80 & 23.98 & 31.06 & 23.13 \\
            & GC-PLM & 36.66 & 37.21 & 37.18 & 42.21 & 55.39 & 41.73 \\
            & DIFT (\textit{Single}) & 40.99 & 40.64 & 39.09 & 44.54 & 61.38 & 45.33 \\
            \cline{2-8}
            & Ours  & \textbf{42.96} & \textbf{42.58} & \textbf{41.69} & \textbf{48.33} & \textbf{63.74} & \textbf{47.86} \\
        \bottomrule
    \end{tabular}
    \caption{
        This table presents the MKGC results across five languages.
        The embedding-based methods TransE \citep{bordes2013translating}, Analogy \citep{10.5555/3305890.3305905}, ComplEx \citep{10.5555/3045390.3045609}, DistMult \citep{Yang2014EmbeddingEA}, and RotatE \citep{sun2018rotate} are all implemented using the OpenKE framework \citep{han2018openke}.
        % TODO add
        The results of HAKE \citep{zhang2020learning} were reproduced using its open-source code.
        ICL refers to evaluation using the LLaMA-2-7b-chat model without fine-tuning.
        GC-PLM \citep{song-etal-2023-multilingual} represents the current SOTA method for MKGC.
        DIFT \citep{liu2024finetuning} is a SOTA LLM-based monolingual KGC method.
        The \textit{Single} refers to training a separate model for each language independently.
        The numbers in bold represent the best results among the methods and languages considered.
    }
    \label{tab:performance}
\end{table}

\section{Experiment}

\subsection{Implementation Details}
In the experiment, we selected TransE \citep{bordes2013translating} to obtain candidate entities.
Additionally, we utilized Llama-2-7b-chat-hf\footnote{\href{https://huggingface.co/meta-llama/Llama-2-7b-chat-hf}{https://huggingface.co/meta-llama/Llama-2-7b-chat-hf}} as the base model for fine-tuning.
The model training hyperparameters are set as follows: the learning rate is 2e-5, the LoRA rank is 4, and the length of the candidate entities \( M_c\) is between 25 and 30.
The number of iterations \(N_t \) for the IER is 10.

\subsection{Multilingual Knowledge Graph Completion}
We compared the performance of the proposed framework with embedding-based and generation-based methods on our constructed dataset.
The experimental results demonstrate that our method achieves optimal performance on the average metrics across all languages.
Specifically, as shown in Table~\ref{tab:performance}, the performance of our proposed framework surpasses all the aforementioned methods in the three languages: EN, FR, and IT.
For JA and ZH, our framework performed excellently on all metrics except Hits@10. 
Our framework failed to surpass RotatE's performance on Hits@10, primarily attributed to our use of a relatively weaker-performing TransE model for generating candidate entities.
We replaced TransE with RotatE in the candidate entities retrieval and conducted experiments. 
The corresponding results and analysis are presented in Appendix~\ref{Appendix:rotate_ours}.
Compared to the existing SOTA MKGC method GC-PLM, our framework achieved significant performance advantages in Hits@1, Hits@3, Hits@10, and MRR metrics, with improvements of 5.47\%, 3.27\%, 1.01\%, and 6.13\%, respectively.
Furthermore, our framework achieves a substantial improvement in performance compared with DIFT, the SOTA LLM-based monolingual KGC method.
Overall, the experimental results clearly demonstrate the effectiveness and superiority of our proposed framework.

\begin{table}[t]
    \centering
    \small
    \setlength{\tabcolsep}{2.5pt}
    \renewcommand{\arraystretch}{1.06}
    \begin{tabular}{p{0.3cm}ccccccc}
        \toprule
         & MODEL & EN & FR & IT & JA & ZH & AVG \\
        \midrule
        \multirow{3}{*}{\shortstack{H \\ @ \\ 1}}
            & LoRAMoE & 36.28 & 36.36 & 35.81 & 40.22 & 56.87 & 41.11 \\
            & HydraLoRA & 35.68 & 35.60 & 35.05 & 40.49 & 57.92 & 40.95 \\
            \cline{2-8}
            & Ours & \textbf{36.50} & \textbf{36.72} & \textbf{35.93} & \textbf{41.60} & \textbf{58.63} & \textbf{41.88} \\
        \midrule
        \multirow{3}{*}{\shortstack{H \\ @ \\ 3}} 
            & LoRAMoE & 42.54 & 42.74 & 41.40 & 48.75 & 64.96 & 48.08 \\
            & HydraLoRA & 42.49 & 42.58 & 41.22 & 48.68 & 64.99 & 47.99 \\
            \cline{2-8}
            & Ours  & \textbf{42.87} & \textbf{43.08} & \textbf{41.70} & \textbf{48.94} & \textbf{65.06} & \textbf{48.33} \\
        \midrule
        \multirow{3}{*}{\shortstack{H \\ @ \\ 10}}
            & LoRAMoE & 52.27 & 52.50 & 50.91 & 59.07 & 72.19 & 57.39 \\
            & HydraLoRA & 52.51 & 52.41 & 50.81 & \textbf{59.10} & 72.24 & 57.41 \\
            \cline{2-8}
            & Ours  & \textbf{52.63} & \textbf{52.62} & \textbf{51.17} & 59.09 & \textbf{72.32} & \textbf{57.57} \\
        \midrule
        \multirow{3}{*}{\shortstack{M \\ R \\ R}}
            & LoRAMoE & 41.15 & 41.13 & 40.30 & 45.90 & 61.52 & 46.00 \\
            & HydraLoRA & 40.80 & 40.62 & 39.81 & 46.06 & 62.18 & 45.89 \\
            \cline{2-8}
            & Ours  & \textbf{41.42} & \textbf{41.44} & \textbf{40.51} & \textbf{46.80} & \textbf{62.67} & \textbf{46.57} \\
        \bottomrule
    \end{tabular}
    \caption{
        This table compares the KL-GMoE with the existing SOTA fine-tuning methods LoRAMoE \citep{dou-etal-2024-loramoe} and HydraLoRA \citep{tian2024hydralora}.
    }
    \label{tab:model-architecture}
\end{table}
\begin{table}[t]
    \centering
    \small
    \setlength{\tabcolsep}{1.2pt}
    \begin{tabular}{lcccccccc}
        \toprule
        Model & Trainable Params & Activated Params & Lora Rank \\
        \midrule
        TransE & 106.1m & 106.1m & - \\
        DIFT(LoRA) & 159.9*5 m & 159.9*5 m & 64 \\
        LoRAMoE & 19.2m & 19.2m & 4 \\
        HydraLoRA & 12.5m & 12.5m & 4 \\
        \midrule
        Ours & 32.9 m & 9.4 m & 4 \\
        \bottomrule
    \end{tabular}
    \caption{
        This table shows the comparison of our method with other methods in terms of parameter count.
        }
    \label{tab:params}
\end{table}
\begin{table}[t]
    \centering
    \begin{tabular}{lcccccccc}
        \toprule
        Model & Avg Tokens Num & TFLOPs \\
        \midrule
        LoRAMoE & 353.89 & 2.37814 \\
        HydraLoRA & 353.89 & 2.37580 \\
        DIFT(LoRA) & 353.89 & 2.42721 \\
        \midrule
        Ours & 353.89 & 2.37472 \\
        \bottomrule
    \end{tabular}
    \caption{
        This table compares the computational efficiency of our method with that of other methods. The data presented are average values calculated from 1,000 samples.
    }
    \label{tab:tflop}
\end{table}

\subsection{Model Architecture Comparison and Parameter Analysis}
We compared the MKGC performance between the KL-GMoE architecture and existing SOTA fine-tuning methods, including LoRAMoE and HydraLoRA.
These two SOTA methods both utilize multiple channels to process a single query, which can lead to the problem of knowledge fragmentation.
As shown in Table~\ref{tab:model-architecture}, KL-GMoE outperforms these methods on average metrics.
This experimental result demonstrates that our method effectively addresses knowledge fragmentation, thereby enhancing performance on the MKGC task.

We further analyzed the advantages of KL-GMoE in terms of model parameters.
As shown in Table \ref{tab:params}, compared to the embedding-based method TransE, KL-GMoE has 3.2 times fewer trainable parameters and 11.3 times fewer activated parameters.
Among LLM-based methods, KL-GMoE has significantly fewer activated parameters than all other methods.
In particular, compared to DIFT, KL-GMoE has approximately 24.3 times fewer trainable parameters and about 85.1 times fewer activated parameters, which demonstrates its significant advantages in terms of parameter count.
At the same time, we compared the proposed method with other methods in terms of FLOPs during inference.
As shown in Table \ref{tab:tflop}, our method reduces the FLOPs by approximately 0.053 TFLOPs compared to the current state-of-the-art LLM-based KGC method, DIFT, demonstrating superior computational efficiency.

\begin{figure}[t]
  \centering
  \includegraphics[width=\linewidth]{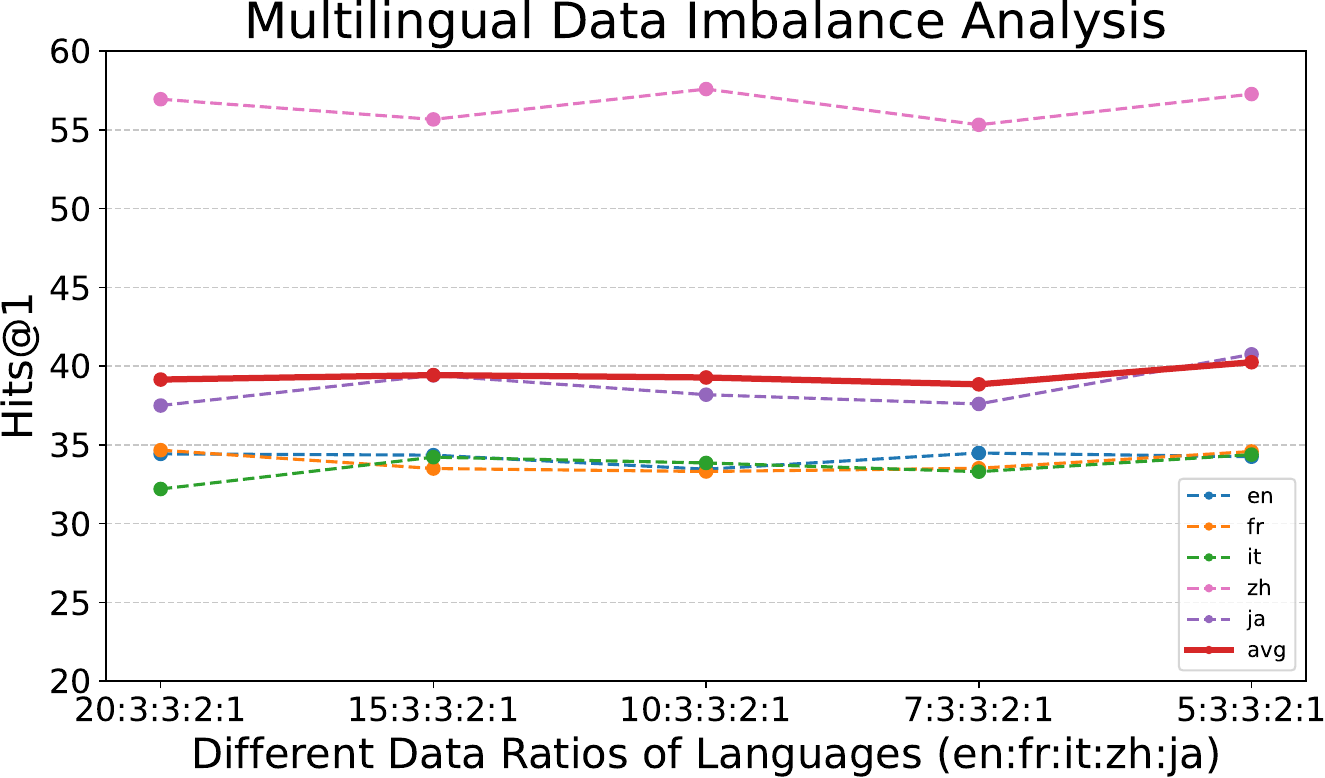}  
  \caption{
    This figure shows the variation in Hits@1 scores of our framework under training data settings with five different language proportions.
  }
  \label{fig:unbalance-hits1-line-all}
\end{figure}

\subsection{Analysis of Language Imbalance}
\definecolor{imbalance-red}{RGB}{197, 58, 50}
To evaluate the robustness of our framework in scenarios with imbalanced language distribution in the training data, we conducted experiments.
Specifically, we conducted experiments with imbalanced training data ratios across five languages, while keeping the total amount of training data constant.
As shown in Figure~\ref{fig:unbalance-hits1-line-all}, despite the changes in language proportions, the Hits@1 scores for each language (dashed lines) and the average score across the five languages (solid \textcolor{imbalance-red}{red} line) remained relatively stable.
It is evident that our framework is insensitive to variations in the language distribution.
Based on this analysis, our framework can effectively leverage cross-lingual shared knowledge, thereby demonstrating strong robustness.

\begin{figure}[t]
  \centering 
  \includegraphics[width=\linewidth]{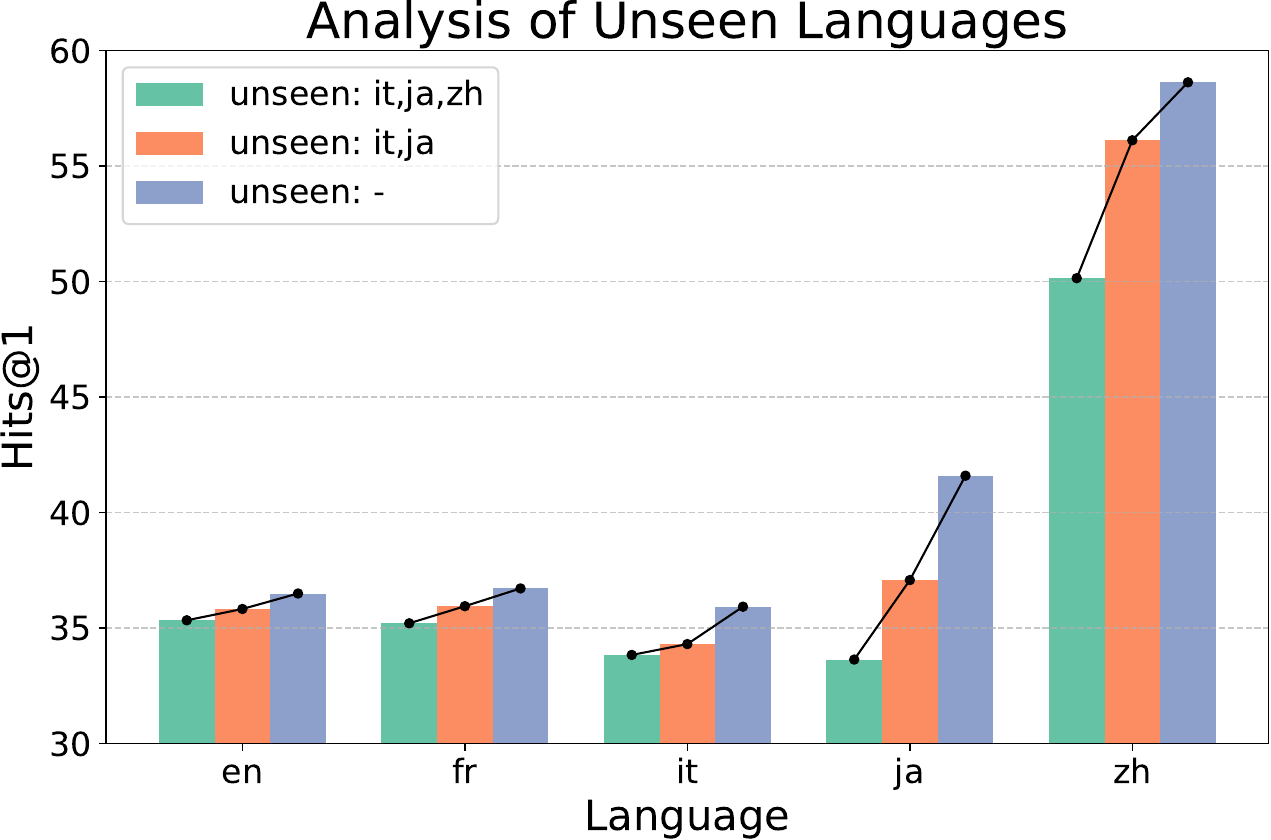} 
  \caption{
    The figure illustrates the Hits@1 performance of our method on five languages under three different training language settings.
  }
  \label{fig:unseen_languages}
\end{figure}

\begin{figure}[t]
  \centering 
  \includegraphics[width=\linewidth]{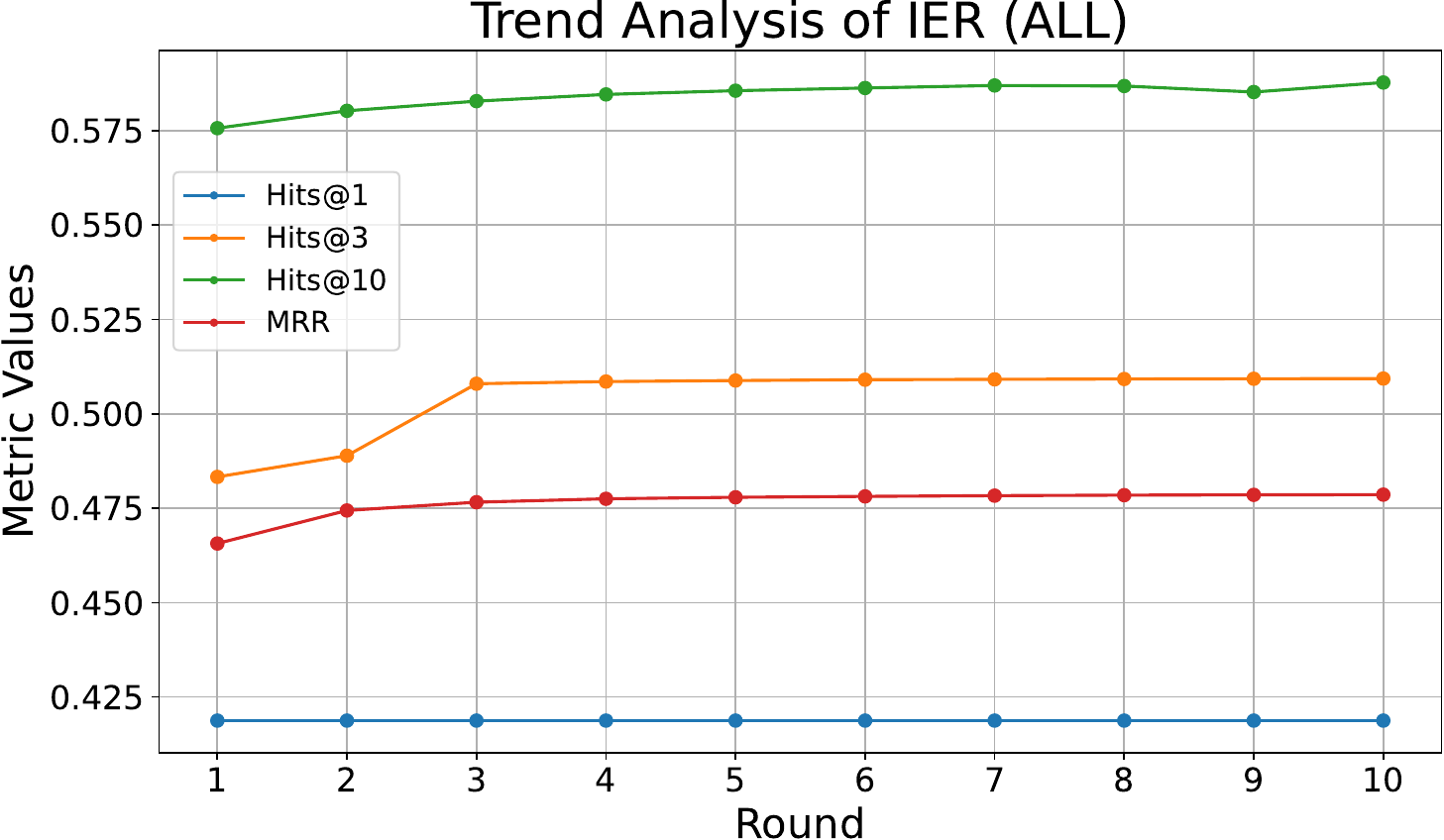}  
  \caption{
    The figure illustrates the impact of the number of iterations of the IER method on performance.
  }
  \label{fig:reranking-trend-avg}
\end{figure}

\begin{figure*}[t]
  \centering
  \includegraphics[width=\linewidth]{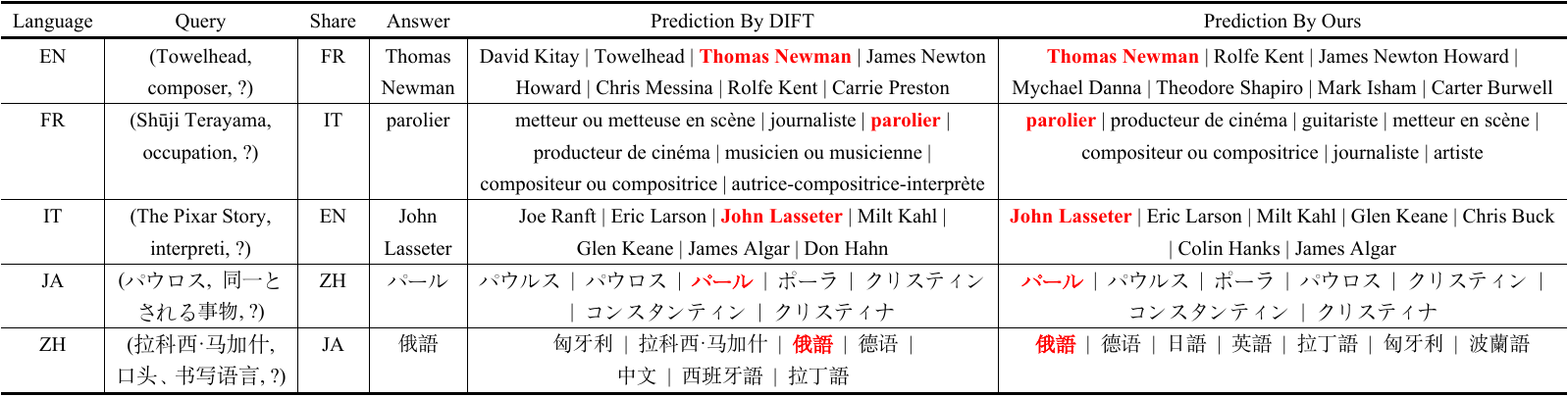}
  \caption{
  The figure presents a comparison of the prediction results between our method and DIFT in the knowledge shared case.
  The \textbf{Share} column indicates that the knowledge of these queries exists in the LLM's training data but is presented in other languages.
  }
  \label{fig:case-study}
\end{figure*}

\subsection{Analysis of Unseen Languages}
\definecolor{unseen-green}{RGB}{125, 192, 166}
To evaluate the generalization capabilities of the proposed framework on languages not included in the training data, we conducted analysis experiments.
These experiments were conducted with three distinct training configurations: (1) trained on EN and FR; (2) trained on EN, FR, and ZH; and (3) trained on five languages.
As shown in Figure~\ref{fig:unseen_languages}, the \textcolor{unseen-green}{green} bar indicates that LLMs trained solely on EN and FR data demonstrated significant KGC performance on unseen languages IT, JA, and ZH.
This clearly demonstrates that knowledge sharing is effective not only among languages seen during LLM training, but also shows significant cross-lingual generalization capability among unseen languages.
Furthermore, we observed a consistent improvement in performance across all languages as the number of training languages increased. 
This finding suggests that training data in more languages provides richer knowledge signals to LLMs, which facilitates the sharing of multilingual knowledge.

\subsection{Analysis of IER Trends}
To evaluate the impact of the number of iterations in the IER method on MKGC performance, we conducted analytical experiments.
Figure~\ref{fig:reranking-trend-avg} illustrates the changes in all metrics as the number of iterations increases.
From the results, it can be observed that Hits@3, Hits@10, and MRR significantly improved in the first three iterations and reached their optimal values by the tenth iteration.
This trend indicates that with an increasing number of iterations, IER allows LLMs to leverage multilingual shared knowledge more effectively, thereby significantly improving the performance of MKGC.

\subsection{Ablation Experiment}
To verify the effectiveness of each component in our proposed framework, we conducted ablation experiment.
We evaluated the contribution of each component by removing it sequentially.
As shown in Table \ref{tab:ablation}, removing the KL-GMoE component resulted in a drop in Hits@1 from 41.88 to 40.28, Hits@3 from 50.93 to 49.71, Hits@10 from 58.78 to 58.07, and MRR from 47.86 to 46.55. 
This indicates that the KL-GMoE component is crucial for improving the performance of MKGC. 
Furthermore, when we removed both KL-GMoE and IER simultaneously, the values of Hits@3, Hits@10, and MRR further decrease compared to removing only KL-GMoE.
This demonstrates that the IER component also makes a positive contribution to the performance of MKGC.
These ablation experiment results strongly prove the effectiveness of our proposed KL-GMoE and IER components.

\begin{table}[t]
    \centering
    % \small 
    \setlength{\tabcolsep}{3.8pt} 
    \begin{tabular}{lcccccccc}
        \toprule
        Model & H@1 & H@3 & H@10 & MRR \\
        \midrule
        Ours & \textbf{41.88} & \textbf{50.93} & \textbf{58.78} & \textbf{47.86} \\
        % \midrule
        Ours w/o \textbf{\textit{kg}} & 40.28 & 49.71 & 58.07 & 46.55 \\
        Ours w/o \textbf{\textit{kg+ier}} & 40.28 & 47.66 & 57.29 & 45.42 \\
        \bottomrule
    \end{tabular}
    \caption{
        This table shows the results of ablation experiments on the KL-GMoE (\textbf{\textit{kg}}) and IER (\textbf{\textit{ier}}) components.
        All results are the average of the five language metrics.
    }
    \label{tab:ablation}
\end{table}

\subsection{Case Study}
We conducted a case study to evaluate the framework's performance in cross-lingual knowledge sharing.
These case' queries are knowledge that the LLM learned during its training, but expressed in another language.
As shown in Figure \ref{fig:case-study}, for the English query {\textit{(Towelhead, composer, ?)}, the LLM has already learned this knowledge in the French training data.
Our framework successfully leverages this French knowledge to accurately predict the entity as {\textit{Thomas Newman}.
In contrast, SOTA LLM-based methods incorrectly predict {\textit{David Kitay}.
This demonstrates that our framework can effectively utilize cross-lingual shared knowledge to improve completion accuracy.

\section{Related Work}
Embedding-based methods map entities and relations in KGs to low-dimensional vector spaces.
For example, TransE \citep{bordes2013translating} based on the translation principle of entities and relations.
RotatE \citep{sun2018rotate} treats each relation as rotation in complex vector space.
DMoG \citep{song-etal-2022-decoupling} represents the unseen relations of the factual graph by fusing ontology and textual graphs.
TransH \citep{10.5555/2893873.2894046} models relation as hyperplane. 
HOLEX \citep{xue2018expanding} interpolate between a high model complexity method and HolE \citep{nickel2016holographic}.
TR-GCN \citep{song2022ontology} proposes an ontology-guided zero-shot relation learning method to represent unseen relations.

Generation-based Methods transform KGC task into text generation task.
For example, KGT5 \citep{saxena-etal-2022-sequence} posing KG link prediction as a sequence-to-sequence task.
GC-PLM \citep{song-etal-2023-multilingual} enhances the performance of MKGC by introducing global and local knowledge constraints.
GenKGC \citep{10.1145/3487553.3524238} introduces a hierarchical decoding strategy of relation-guided demonstration and entity awareness.
KICGPT \citep{wei-etal-2023-kicgpt} integrates LLMs and KGE model, adopting a knowledge-prompted contextual learning strategy to rerank multiple entities.
DIFT \citep{liu2024finetuning} fine-tunes LLMs using LoRA \citep{hu2022lora} to select the most optimal entity from candidate entities obtained by the KGE model.

\section{Conclusion}

In this paper, we propose a novel MKGC framework.
This framework integrates two components: KL-GMoE and IER.
KL-GMoE significantly improves completion performance by efficiently capturing shared knowledge across languages.
IER fully utilized cross-lingual shared knowledge through a multi-round iterative approach, further improving completion performance.
The experimental results demonstrate that our framework exhibits superior performance in the MKGC task.

\section*{Limitations}
Our framework is limited by the token length of the LLM, therefore it is unable to perform entity selection based on all entities in the KG.
Moreover, the framework processes text information exclusively. 
This limitation impedes its application to multimodal KG datasets, as it cannot integrate information from other modalities.

\section*{Ethics Statement}
The paper proposes a method for MKGC and conducts experiments on a multilingual dataset extended from public available datasets.
Therefore, data privacy implications are non-existent in this scenario.

\section*{Acknowledgements}
This research was supported by the National Natural Science Foundation of China (Grant Nos. U21B2027, U23A2038, 62166023, 62376270), the Yunnan Provincial Major Science and Technology Special Plan Projects (Grant Nos. 202402AG050007, 202502AD080012, 202502AD080016), the General Projects of Basic Research in Yunnan Province (Grant Nos. 202301AS070047, 202201BE070001-021).

\bibliography{anthology,custom}
\bibliographystyle{acl_natbib}

\appendix
% \parewpage
\clearpage

\onecolumn
\section{Appendix}

\subsection{Prompt Example}
\label{Appendix:prompt-example}
We present prompt examples for candidate entity lists of varying lengths during the training phase.
\begin{table*}[!ht]
    \centering
    \renewcommand{\arraystretch}{1.1} % 设置行间距
    \begin{tabular}{p{1.5cm}p{13cm}}
        % \toprule
        % Triplet &  (Ken Ogata, award received, Japan Academy Prize) \\
        \midrule
        Prompt & Given a triplet with a missing tail entity t: (Saint George and the Dragon, material used, t).\par\vspace{0.6em}\par
        The following provides descriptive information about entity Saint George and the Dragon:\par
        Saint George and the Dragon, Saint George and the Dragon or Saint George Killing the Dragon is a 1555 or 1558 painting by the Venetian artist Tintoretto. It was later acquired by the English collector\par\vspace{0.6em}\par
        Here are some triplets containing entity Saint George and the Dragon:\par
        [(Saint George and the Dragon, depicts, hill); (Saint George and the Dragon, depicts, spear); (Saint George and the Dragon, creator, Jacopo Tintoretto); (Saint George and the Dragon, depicts, combat); (Saint George and the Dragon, depicts, woman); (Saint George and the Dragon, depicts, sky)]\par\vspace{0.6em}\par
        What is the entity name of t? Select one from the list of entities below: [oil paint; Saint George and the Dragon; wood; tempera; textile; brick; pearl; metamorphic rock; schist; sandstone; paint; igneous rock; tissue; gemstone; brass; copper; woven fabric; volcanic rock; marble; dragon; basalt; sedimentary rock; The Three Graces; limestone; steel]\par\vspace{0.6em}\par
        [Answer]:  \\
        \midrule
        Number of entities & 25 \\
        \midrule
        \midrule
        Prompt & Given a triplet with a missing tail entity t: (Jason Lee, instance of, t).\par\vspace{0.6em}\par
        The following provides descriptive information about entity Jason Lee:\par
        Jason Lee, Jason Michael Lee (born April 25, 1970) is an American actor, photographer, producer, skateboarder, comedian, and writer. He is best known for his roles as Earl Hickey in the television\par\vspace{0.6em}\par
        Here are some triplets containing entity Jason Lee:\par
        [(Mallrats, cast member, Jason Lee); (Jason Lee, ethnic group, Scottish American); (Jason Lee, occupation, screenwriter); (Jason Lee, occupation, actor); (Jason Lee, occupation, film producer); (Jason Lee, occupation, businessperson)]\par\vspace{0.6em}\par
        What is the entity name of t? Select one from the list of entities below: [Jason Lee; human; twin; Jason Alexander; Sofía Vergara; Kevin Smith; Screen Actors Guild Award; David Cross; 3D film; college; Primetime Emmy Award; sports season; MTV Movie Awards; Kaley Cuoco; municipality of Spain; Jason Mewes; decade; military rank; suburb; animation studio; Jane Lynch; Hank Azaria; Satellite Award; Breckin Meyer; My Name Is Earl; Patrick Warburton; business]\par\vspace{0.6em}\par
        [Answer]:  \\
        \midrule
        Number of entities & 27 \\
        \bottomrule
    \end{tabular}
    \caption{
        Prompt examples for candidate entity lists of varying lengths.}
    \label{tab:prompt-example}
\end{table*}

\subsection{Details of the Iterative Entity Reranking Algorithm}
\label{Appendix:ier-alg}

% \begin{algorithm}[h]
% \caption{Iterative Entity Reranking (IER)}
% \label{sec:reranking}
% \begin{algorithmic}[1]
% \STATE Given a query \( q = (h, r, ?)\), where \( ? \) denotes an entity that needs to be predicted.
% \STATE Use the KGE model to score all entities, generating an ordered list \( M = [e_1, e_2, \dots, e_m, \dots, e_n] \).
% \STATE Select the top \( m \) entities from \( M \) to form the candidate entities \( M_c^{(1)} = [e_1, e_2, \dots, e_m] \).
% \STATE Set \( N_t = 10 \).
% \FOR{\( t = 1 \) to \( N_t \)}
%     \STATE Use LLM to select the most relevant \( e^{(t)} \in M_c^{(t)} \) to complete \( q \).
%     \STATE \( M_c^{(t+1)} = M_c^{(t)} \setminus e^{(t)} \)
%     % \STATE Adjust the rank of entity \( e^{(t)} \) in \( M_c^{(1)} \) to \( t \).
%     \STATE Remove \( e^{(t)} \) from \( M_c^{(1)} \).
%     \STATE Insert \( e^{(t)} \) into the \(t\)-th position in \( M_c^{(1)} \).
% \ENDFOR
% \STATE \textbf{Return:}  \( M_c^{(1)} \).
% \end{algorithmic}
% \end{algorithm}

\begin{algorithm}[h]
\caption{Iterative Entity Reranking (IER)}
\label{sec:ier}
\begin{algorithmic}[1]
\STATE \textbf{Input:} Query \( q = (h, r, ?) \), \( M_c^{(1)} = [e_1, e_2, \dots, e_m] \): the top-\( m \) entities generated by the KGE model, \( N_t \), \( L^{(1)}=M_c^{(1)} \)
\FOR{$t = 1$ to $N_t$}
    % \STATE Use LLM to select the most relevant \( e^{(t)} \in M_c^{(t)} \) to complete \( q \).
    \STATE \( e^{(t)} = \underset{e_i \in M_c^{(t)}}{\text{argmax}} \, P(e_i \mid h, r, M_c^{(t)}) \);
    \STATE \( M_c^{(t+1)} = M_c^{(t)} \setminus \{ e^{(t)} \} \);
    \STATE \( L^{(t+1)} = \textbf{Insert}(L^{(t)} \setminus \{e^{(t)}\}, t, e^{(t)}) \);
    % \STATE Remove \( e^{(t)} \) from \( L^{(t)} \).
    % \STATE Insert \( e^{(t)} \) into the \(t\)-th position in \(L^{(t)} \).
\ENDFOR
\STATE \textbf{Output:} \( L^{(N_t+1)} \)
\end{algorithmic}
\end{algorithm}

% TODO add
\subsection{Analysis of Knowledge Graph Embedding Models}
\label{Appendix:rotate_ours}
The experimental results in Table ~\ref{tab:kge_analysis} clearly demonstrate that when using RotatE to retrieve candidate entities, our proposed method achieves a significant performance improvement compared to the original RotatE model, with a 14.78\% increase in Hits@1.
Notably, Ours+RotatE exhibits slightly lower performance than Ours+TransE on several language-specific metrics.
This phenomenon can be attributed to the differing top-1 ranking rates of correct entities within the candidate sets generated by each KGE model.
Specifically, the proportion of correct entities ranked as top-1 was 14.38\% when using TransE, whereas this proportion significantly increased to 30.51\% with RotatE. 
Therefore, this feature has had some impact: during the fine-tuning stage, LLM is more inclined to choose the entity that ranks first in the candidate set as the final answer. 
We hypothesize that this "top-1 bias" may, to some extent, suppress the model's exploration of other potentially correct answers, leading to Ours+RotatE performing slightly worse than Ours+TransE on some languages.
In future work, we plan to further investigate how to construct a more stable fine-tuning instruction set that does not rely on traditional KGE models.
\begin{table}[h]
    \centering
    \begin{tabular}{p{0.8cm}ccccccc}
        \toprule
         & MODEL & EN & FR & IT & JA & ZH & AVG \\
        \midrule
        \multirow{3}{*}{\shortstack{H \\ @ \\ 1}}
            & RotatE & 24.08 & 24.61 & 25.57 & 29.49 & 31.36 & 27.02  \\
            & Ours+TransE & 36.50 & \textbf{36.72} & \textbf{35.93} & 41.60 & 58.63 & \textbf{41.88} \\
            & Ours+RotatE & \textbf{36.55} & 35.70 & 35.46 & \textbf{41.78} & \textbf{59.49} & 41.80 \\
        \midrule
        \multirow{3}{*}{\shortstack{H \\ @ \\ 3}} 
            & RotatE & 40.73 & 42.13 & 41.68 & 49.78 & 62.36 & 47.34  \\
            & Ours+TransE  & 46.25 & \textbf{45.30} & \textbf{44.22} & 51.97 & 66.93 & 50.93 \\
            & Ours+RotatE & \textbf{46.38} & 45.05 & 44.20 & \textbf{52.27} & \textbf{67.13} & \textbf{51.01} \\
        \midrule
        \multirow{3}{*}{\shortstack{H \\ @ \\ 10}}
            & RotatE & 52.66 & 53.17 & 51.50 & 61.68 & 74.58 & 58.72  \\
            & Ours+TransE  & 54.71 & 53.45 & 52.31 & 60.85 & 72.56 & 58.78 \\
            & Ours+RotatE & \textbf{54.76} & \textbf{54.13} & \textbf{52.53} & \textbf{62.31} & \textbf{74.59} & \textbf{59.66} \\
        \midrule
        \multirow{3}{*}{\shortstack{M \\ R \\ R}}
            & RotatE & 34.63 & 35.39 & 35.49 & 41.75 & 48.74 & 39.20 \\
            & Ours+TransE  & 42.96 & \textbf{42.58} & \textbf{41.69} & 48.33 & 63.74 & 47.86 \\
            & Ours+RotatE & \textbf{43.04} & 42.18 & 41.58 & \textbf{49.01} & \textbf{64.46} & \textbf{48.05} \\
        \bottomrule
    \end{tabular}
    \caption{
        The impact of different KGE models on the performance of our proposed framework during the candidate entities retrieval process.
    }
    \label{tab:kge_analysis}
\end{table}

\subsection{Analysis of Expert Routing}
\begin{comment}
为了验证跨语言知识共享的存在，我们对测试样本中的专家选择进行了分析。
我们获取了每个专家路由的输出，并从知识和语言两个维度进行可视化。
图~\ref{fig:route-analyse}的左侧展示了基于语言维度的专家路由分析。
说明了在不同语言的样本中，LLM的每个Transformer层中专家选择的情况。
不同语言样本中的专家选择大多是一致的，这表明我们的方法在MKGC任务中不会区分语言。
图的右侧展示了每个关系的专家选择。
图中的分析表明，不同语言之间具有相同关系的样本大多由相同的专家处理。
总体来看，部分关系选择的专家选择一致，而部分则存在差异。
基于这一分析，可以验证our method能够有效地利用不同语言之间共享知识。
\end{comment}
To verify the existence of knowledge sharing, we analyzed the expert selection in the test samples.
We obtained the output of each expert routing and visualized it from both the linguistic and knowledge dimensions.
The left side of Figure ~\ref{fig:route-analyse} shows the expert routing analysis based on the linguistic dimension. 
Illustrates the selection of experts in each Transformer layer of the LLM for samples in different languages.
The expert selections across different language samples are mostly consistent, suggesting that our method does not distinguish between languages in MKGC task.
The right side of the figure illustrates the expert selection for each relation.
The analysis from figure shows that samples with same relations across different languages are mostly handled by same experts.
Overall, some relations exhibit consistent expert selection, while others show differences.
Based on this analysis, it is validated that our method can effectively leverage knowledge sharing across different languages.

\begin{figure*}[h]
  \centering
  \includegraphics[width=\linewidth]{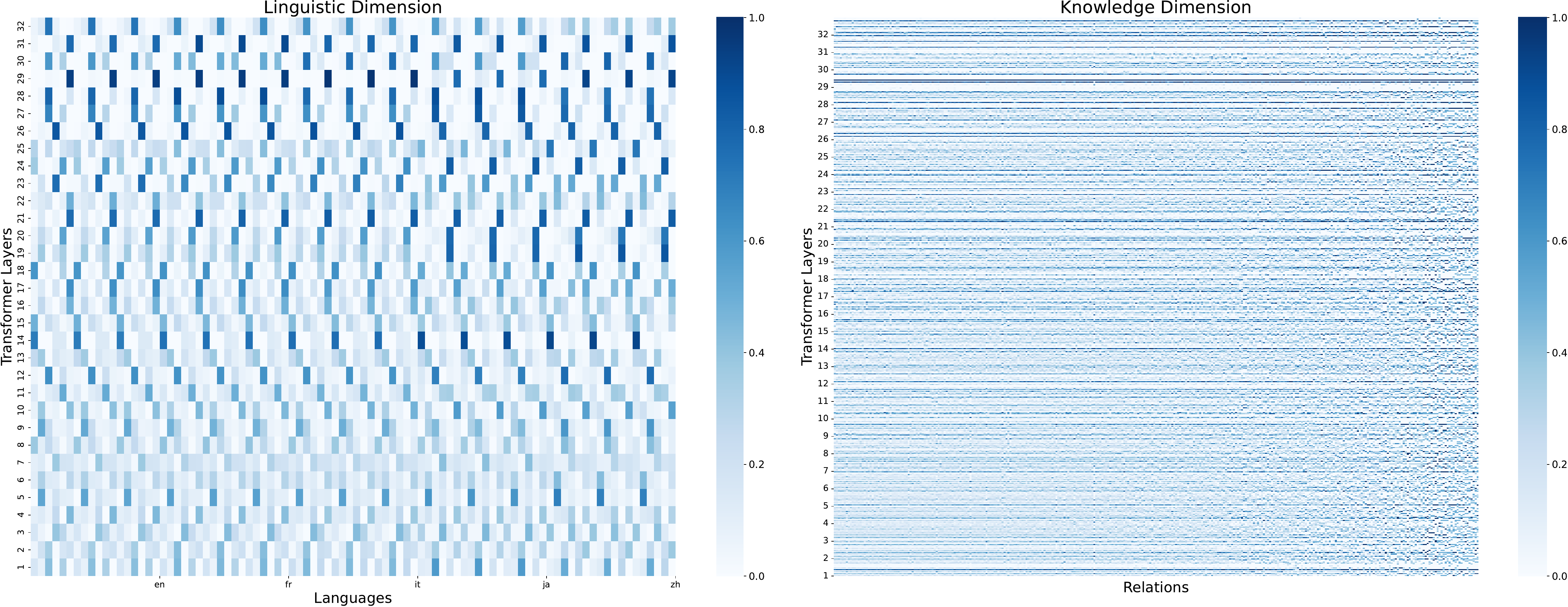} 
  % % 该图展示了测试样本在语言和知识维度上的专家选择情况
  % 左侧显示了在五种语言中的专家选择。
  % 横轴表示所有语言，语言中的每个小条对应一个专家。
  % 纵轴表示LLM中Transformer的层号。  
  % 每个区块的颜色强度表示选择特定专家的样本频率。  
  % 右侧展示了在所有关系中的专家选择。  
  % 横轴表示所有关系，纵轴表示Transformer的层号，层中每一行对应一个专家。
  \caption{
  The left shows expert selection across five languages. 
  The horizontal axis represents all languages, with each small bar within a language corresponding to an expert. 
  The vertical axis indicates the layer numbers in the Transformer of the LLM.
  The color intensity of each blocks represents the frequency of samples selecting particular expert.
  The right depicts expert selection across all relations.
  The horizontal axis represents all relations.
  Vertical axis shows the layer numbers in Transformer, each row within a layer corresponding to an expert.
  }
  \label{fig:route-analyse}
\end{figure*}

\subsection{IER Trend Analysis}
We analyzed the impact of the number of iterations in the IER method on five language evaluation metrics.
The experimental results reveal a clear performance improvement across all languages with increased iterations. 
These findings demonstrate the IER method's ability to exploit cross-lingual shared knowledge.

\twocolumn % 切换为双栏

\begin{figure}[htb!]
  \centering
  \includegraphics[width=\linewidth]{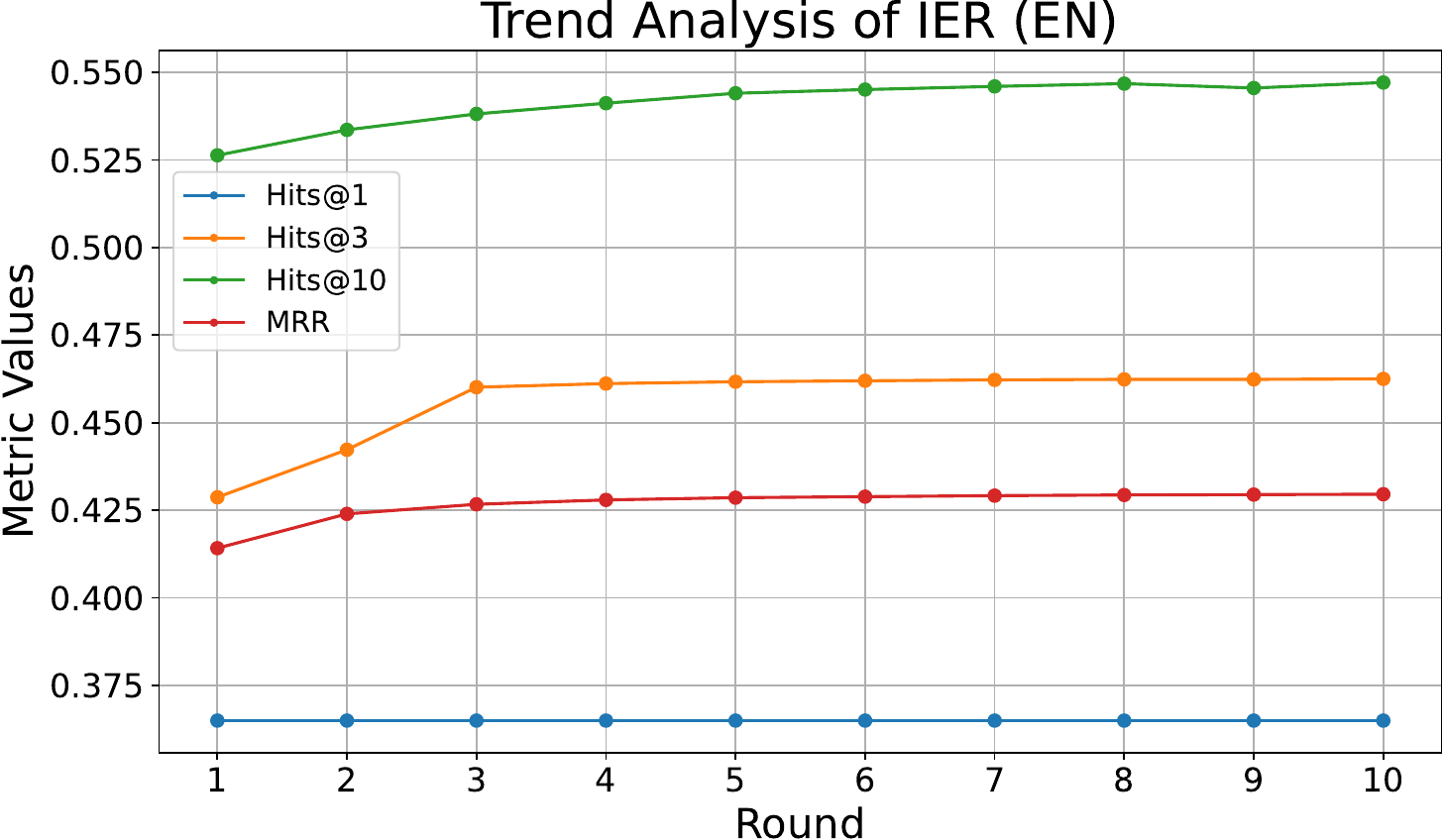}
  \caption{
        This figure shows how the metric performance of the IER method changes with the number of iterations on the English test set.
  }
  \label{fig:reranking-trend-en}
\end{figure}

\begin{figure}[htb!]
  \centering
  \includegraphics[width=\linewidth]{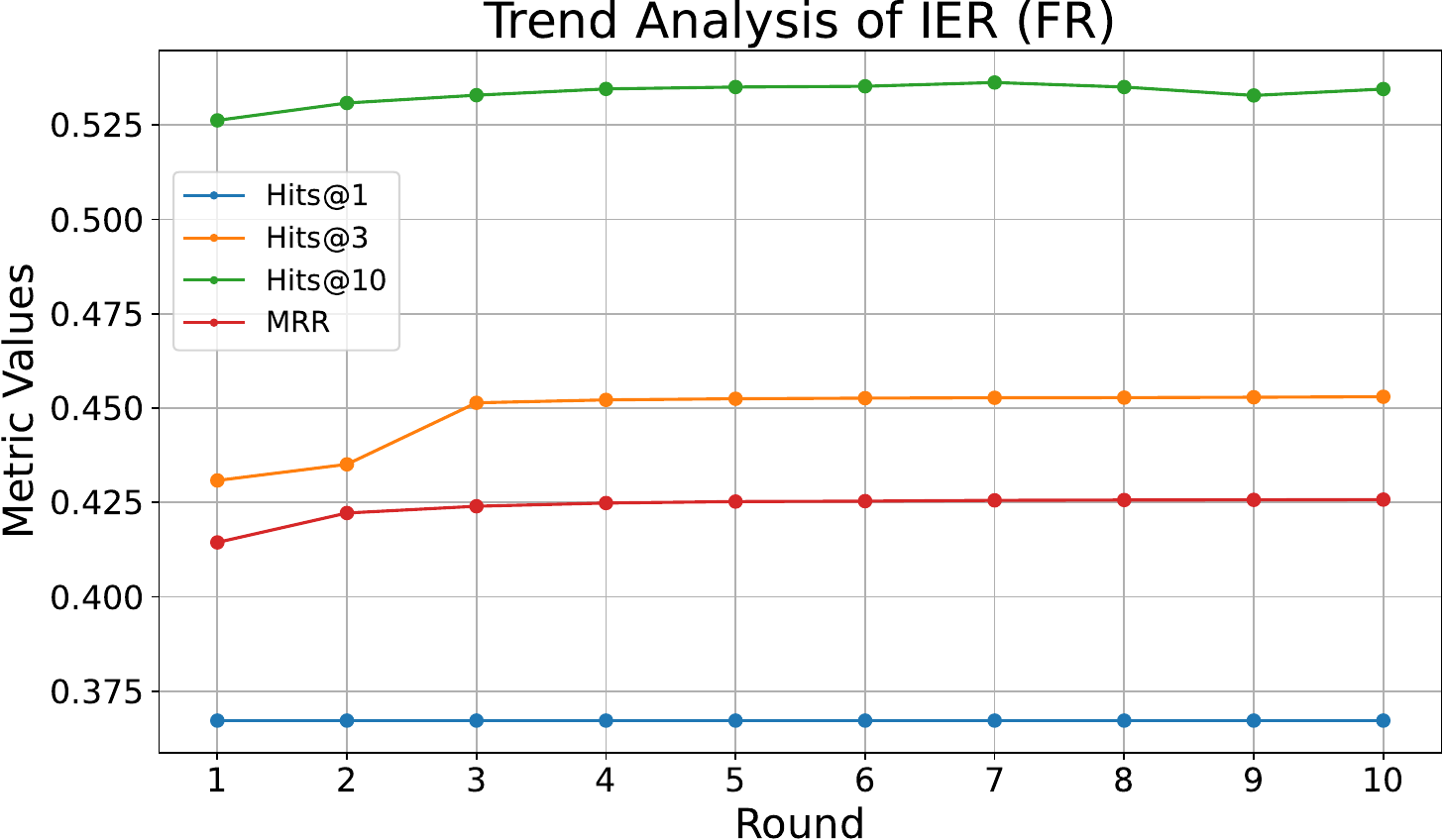}
  \caption{ 
     This figure shows how the metric performance of the IER method changes with the number of iterations on the French test set.
  }
  \label{fig:reranking-trend-fr}
\end{figure}

\begin{figure}[htb!]
  \centering
  \includegraphics[width=\linewidth]{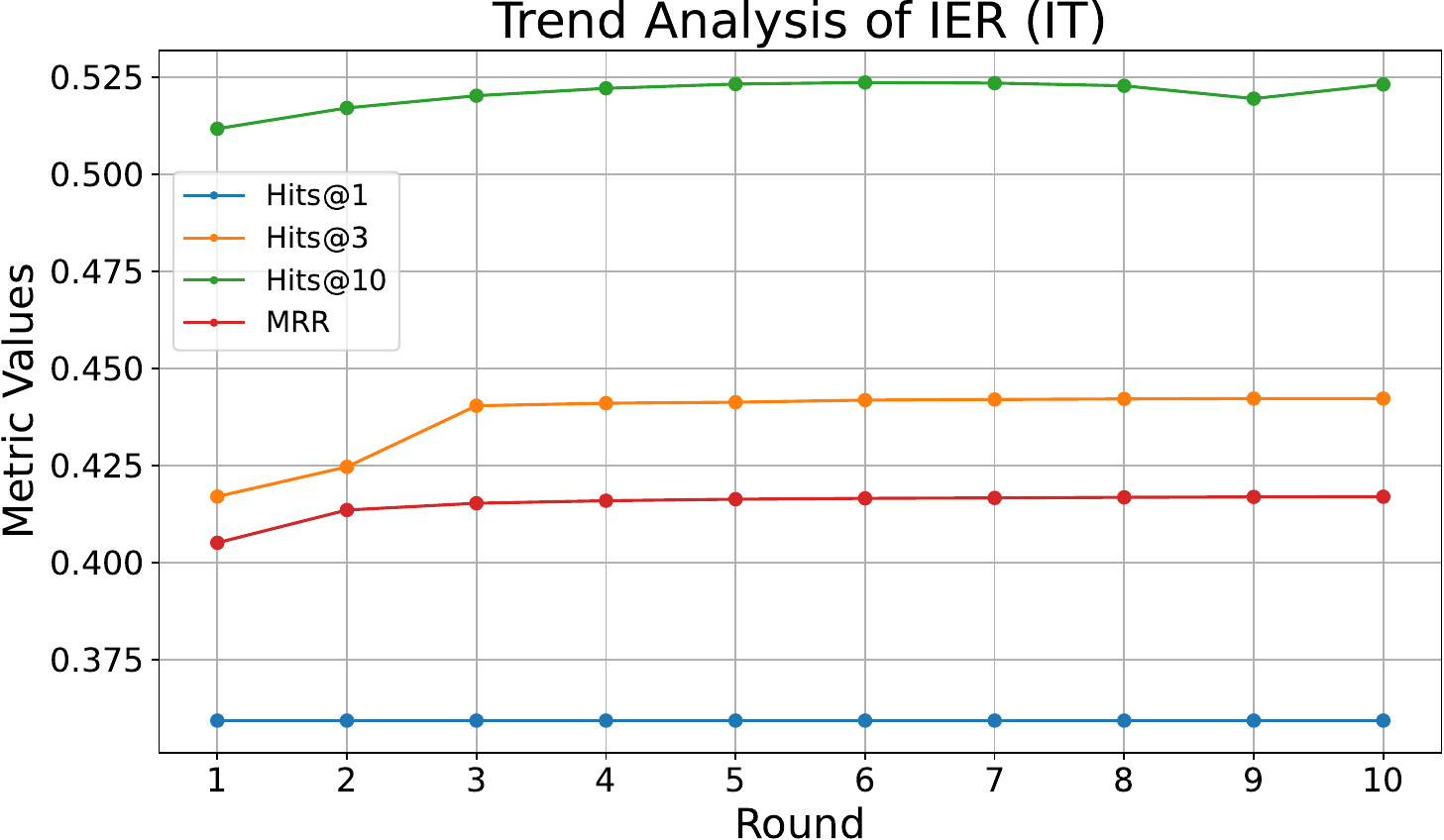}
  \caption{ 
      This figure shows how the metric performance of the IER method changes with the number of iterations on the Italian test set.
  }
  \label{fig:reranking-trend-it}
\end{figure}

\begin{figure}[hbt!]
  \centering
  \includegraphics[width=\linewidth]{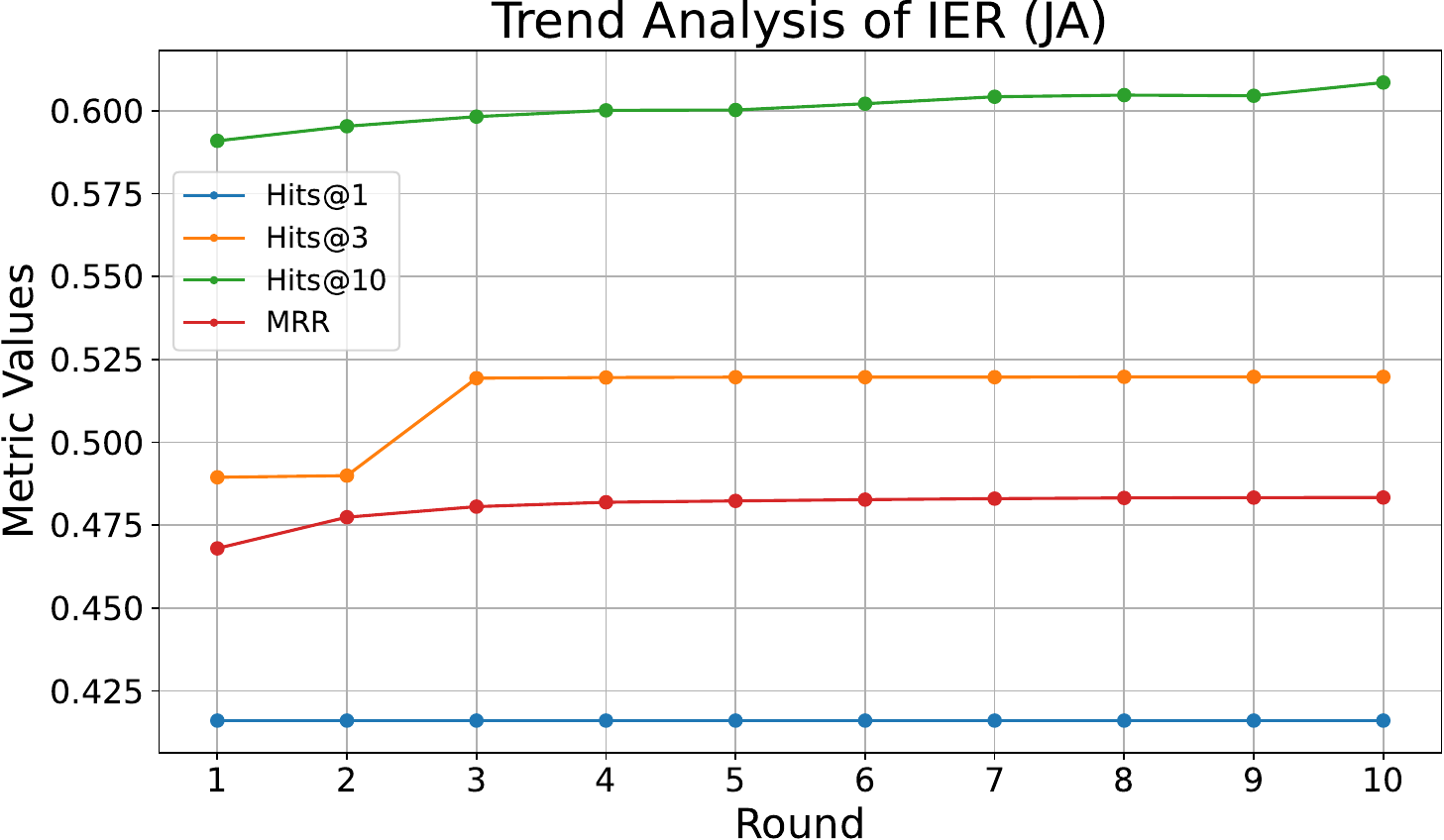}
  \caption{ 
    This figure shows how the metric performance of the IER method changes with the number of iterations on the Japanese test set.
  }
  \label{fig:reranking-trend-ja}
\end{figure}

\begin{figure}[hbt!]
  \centering
  \includegraphics[width=\linewidth]{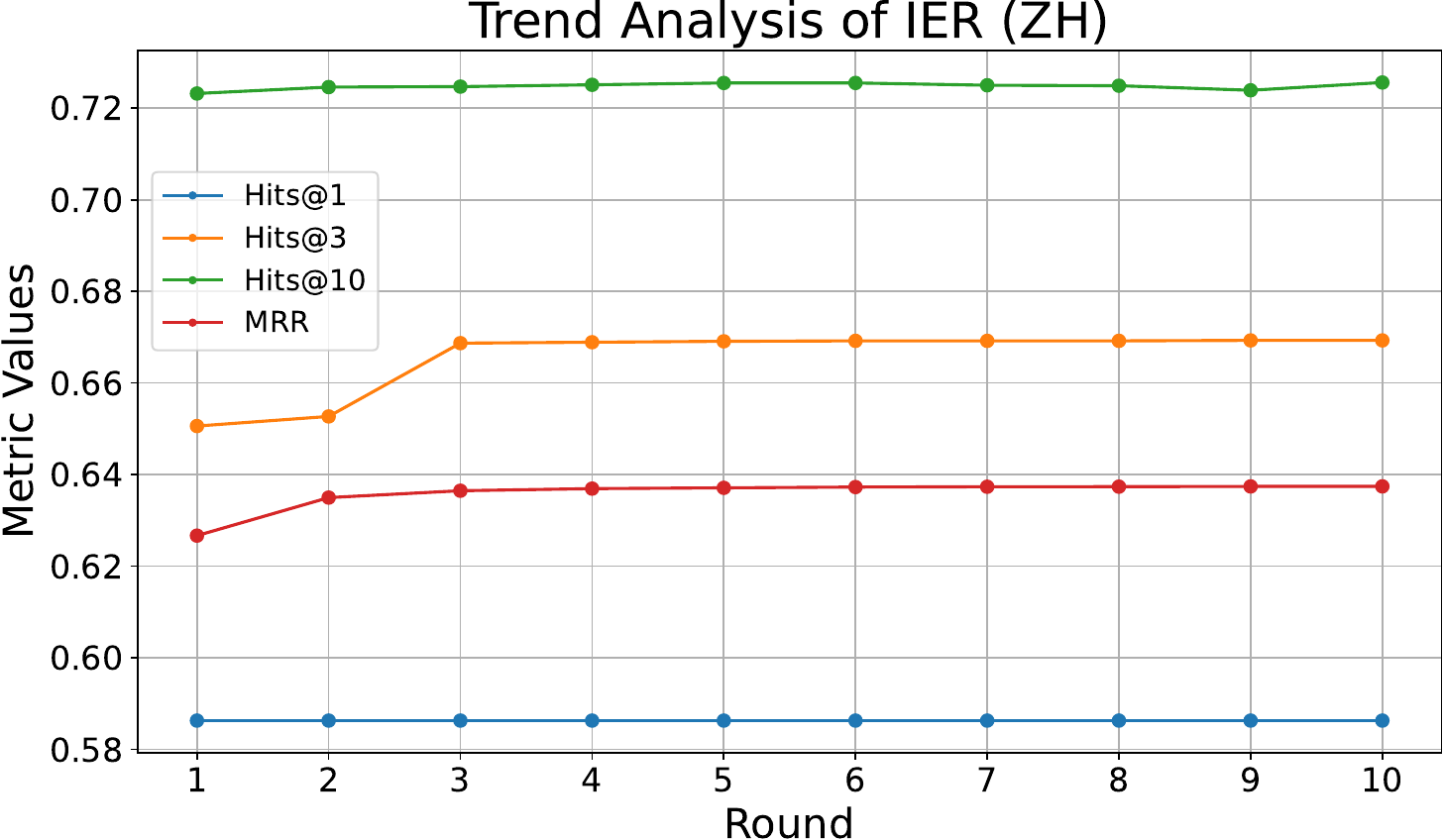}
  \caption{ 
    This figure shows how the metric performance of the IER method changes with the number of iterations on the Chinese test set.
  }
  \label{fig:reranking-trend-zh}
\end{figure}

% \onecolumn
% \subsection{Case Study}
% We demonstrate the significant advantages of our framework over SOTA LLM-based methods for MKGC through a set of case studies.
% The language used in the queries for these case studies differs from the language of the relevant knowledge in the LLM's training data.
% As shown in Figure~\ref{fig:case-study}, for the English query \textbf{\textit{(Towelhead, composer, ?)}}, the knowledge about the correct answer exists in French within the LLM's training data.
% Our framework successfully leverages this French knowledge to accurately predict the entity as \textbf{\textit{Thomas Newman}}.
% In contrast, SOTA methods incorrectly predict \textbf{\textit{David Kitay}}.
% This powerfully demonstrates that our framework can effectively utilize cross-lingual shared knowledge to complete queries.

% \begin{figure*}[h]
%   \centering
%   \includegraphics[width=\linewidth]{pdf/case-study.pdf}
%   \caption{
%   The figure presents a comparison of the prediction results between our method and DIFT in the knowledge shared case.
%   The \textbf{Share} column represents the occurrence of these queries in training set of other languages.
%   }
%   \label{fig:case-study}
% \end{figure*}

\end{document}